%% file: root.tex
\documentclass[letterpaper, 10 pt, conference]{ieeeconf}  

\IEEEoverridecommandlockouts                              
\title{\LARGE \bf
Accelerated Spline-Based Time-Optimal Motion Planning with \\
Continuous Safety Guarantees for Non-Differentially Flat Systems
}

\author{Dries Dirckx and Jan Swevers and Wilm Decr\'{e}
\thanks{This work is funded by Flanders Make through the SBO project ARENA: Agile and
Reliable Navigation and the SBO project LearnOpTra: Learning meets optimization for robust and multimodal trajectory planning}
\thanks{D. Dirckx, J. Swevers and W. Decr\'{e} are with Department of Mechanical Engineering and Flanders Make @ KU Leuven, Faculty of Engineering Science,
        KU Leuven, 3001 Heverlee, Belgium. Corresponding author:
        {\tt\small dries.dirckx@kuleuven.be}}%
}

\input{preamble.tex}

\begin{document}

\maketitle
\thispagestyle{empty}
\pagestyle{empty}

\begin{abstract}

Generating time-optimal, collision-free trajectories for autonomous mobile robots involves a fundamental trade-off between guaranteeing safety and managing computational complexity. State-of-the-art approaches formulate spline-based motion planning as a single Optimal Control Problem (OCP) but often suffer from high computational cost because they include separating hyperplane parameters as decision variables to enforce continuous collision avoidance. This paper presents a novel method that alleviates this bottleneck by decoupling the determination of separating hyperplanes from the OCP. By treating the separation theorem as an independent classification problem solvable via a linear system or quadratic program, the proposed method eliminates hyperplane parameters from the optimisation variables, effectively transforming non-convex constraints into linear ones. Experimental validation demonstrates that this decoupled approach reduces trajectory computation times up to almost 60\% compared to fully coupled methods in obstacle-rich environments, while maintaining rigorous continuous safety guarantees.



\end{abstract}

\section{Introduction: Autonomous Motion Planning}
\input{Sections/introduction}

\section{Related Work}
\input{Sections/related_work}

\section{Spline-based Motion Planning}
\input{Sections/splines}
\section{Accelerated Continuous Collision Avoidance}
\input{Sections/bilevel}

\section{Experimental Validation}
\input{Sections/experiments}

\section{Conclusion and Future Work}
\input{Sections/conclusion}
\printbibliography






\end{document}

%% file: preamble.tex
\usepackage{textcomp} 
\usepackage{pifont}   
\usepackage{booktabs}
\usepackage{amssymb, amsfonts} 
\usepackage{amsmath}
\usepackage{multicol}
\usepackage{graphicx}
\usepackage{textcomp}
\usepackage{xcolor}
\usepackage[utf8]{inputenc}
\usepackage{pgfplots}
\usepackage{pdfpages}
\usepackage{tabularx}
\usepackage{tikz}
\usepackage[font=normalsize]{caption}
\usepackage{tgadventor}
\usetikzlibrary{positioning}
\usepackage{svg}
\usepackage{arydshln}
\usepackage{tcolorbox}
\tcbuselibrary{theorems}
\usepackage{algorithm2e}
\usepackage{gensymb}
\usepackage{tikz-cd}

\definecolor{lightgraybox}{rgb}{0.85, 0.85, 0.85}

\newcolumntype{Y}{>{\centering\arraybackslash}X}

\usepackage[hyperref=auto, mincrossrefs=999, backend=biber, style=numeric-comp]{biblatex}
\usepackage{mathtools}
\usepackage{xurl}  

\usepackage{makecell}
\usepackage{multirow}
\usepackage{optidef}
\usepackage{tikz-cd}

\newcommand{\vect}[1]{\boldsymbol{#1}}

\newcommand{\tablefontsize}{\normalsize}

\makeatletter
\newcommand*\mysize{%
  \@setfontsize\mysize{7}{14}%
}
\makeatother

\usepackage[english]{babel}
\usepackage{amsthm}

\theoremstyle{definition}

\newcolumntype{Y}{>{\centering\arraybackslash}X}
\newcolumntype{P}[1]{>{\centering\arraybackslash}p{#1}}

\newcommand\Set[2]{\{\,#1\mid#2\,\}}

\usepackage{array}
\usepackage{stfloats}
\usepackage{float}
\usepackage{verbatim}
\usepackage[T1]{fontenc}
\usepackage{subcaption}
\usepackage{bm}
\usetikzlibrary{shapes}
\usepackage[export]{adjustbox}
\usepackage{contour}
\usepackage{rotating}
\usepackage[normalem]{ulem}
\newcommand{\green}[1]{{\color{green}#1}}

\DeclareRobustCommand\sampleline[1]{%
  \tikz\draw[#1] (0,0) (0,\the\dimexpr\fontdimen22\textfont2)
  -- (1.5em,\the\dimexpr\fontdimen22\textfont2);%
}

\newcommand{\splitline}[4]{
\tikz{\draw[#1, #3, #4] (0, 0) (0,  \the\dimexpr\fontdimen22\textfont2) -- (0.75em,  \the\dimexpr\fontdimen22\textfont2) edge[#2] +(0:0.75em);}
}

\newtcbtheorem[number within=section]{mytheo}{}%
{colback=lightgray!15,colframe=gray!85, separator sign none}{th}

\definecolor{green}{RGB}{0, 170, 0}
\definecolor{red}{RGB}{200, 0, 0}
\definecolor{lightgray}{RGB}{200, 200, 200}
\definecolor{darkgray}{RGB}{150, 150, 150}
\definecolor{lightblue}{RGB}{41, 200, 200}
\definecolor{darkblue}{RGB}{41, 92, 207}
\definecolor{jump_red}{RGB}{176, 33, 33}
\definecolor{jump_yellow}{RGB}{250, 191, 5}
\definecolor{steelblue}{RGB}{70, 130, 180}
\definecolor{firebrick}{RGB}{178, 34, 34}

\contourlength{0.8pt}

\newcolumntype{L}[1]{>{\raggedright\let\newline\\\arraybackslash\hspace{0pt}}p{#1}}
\newcolumntype{C}[1]{>{\centering\let\newline\\\arraybackslash\hspace{0pt}}p{#1}}
\newcolumntype{R}[1]{>{\raggedleft\let\newline\\\arraybackslash\hspace{0pt}}p{#1}}

%% file: Sections/introduction.tex
\label{sec:introduction}


Autonomous Mobile Robots (AMRs) have become indispensable in modern industrial domains, ranging from material transport in workshops and e-commerce warehouses to agricultural applications. Currently, safety is assured by either separating robots and humans in dedicated tracks or imposing tight limits the forward velocity. To further their applicability and ease their deployment, these robots have to navigate vast, dynamic, and uncertain spaces. The primary challenge herein lies in generating trajectories that are not only cost-efficient, e.g., time-optimal, and physically feasible but also guaranteed to be safe at every time instant. This brings about a fundamental trade-off in motion planning between guaranteeing safety, optimality and the computational cost in finding such a trajectory. \\

Traditionally, autonomous navigation is tackled through a decoupled approach, splitting the task into geometric path planning (ignoring system dynamics) followed by trajectory tracking~\cite{Debrouwere2013, Foehn2021}. While computationally efficient, this separation often yields suboptimal results or infeasible paths because the planner neglects the vehicle's kinodynamic limits imposed by the tracking controller. Conversely, recent coupled approaches formulate motion planning as an optimisation problem~\cite{Romero2022, Schoels2020}, yielding better performance and formal safety guarantees but at a higher computational complexity. This paper aims to alleviate the computational load for a specific class of those methods, spline-based motion planners, while maintaining guarantees on continuous collision avoidance.

%% file: Sections/related_work.tex
\label{sec:related_work}

Infinite-dimensional OCPs are usually transcribed into a finite-dimensional Nonlinear Program (NLP) through time-gridding methods, such as multiple shooting, single shooting or direct collocation~\cite{diehl2011numerical}. They however often suffer from a trade-off between computational cost and safety. Coarse grids allow inter-interval safety violation but are cheap to solve, while fine grids result in computational loads that real-time applicability. Schulman et al.~\cite{Schulman2014} address that trade-off for a path planning algorithm by penalising the signed distance of the swept volume of the robot's convex hull to the obstacles with an exact $\ell_{1}$-penalty. 
Sundaralingam et al.~\cite{curobo} implement a similar but parallellisable procedure sweeping spheres over a single interval, allowing to guarantee continuous collision avoidance with respect to neural-network or voxel world representations. These methods either form a conservative approximation of the rotation of the robot over a single interval or limit the geometric representation of the robot to spheres.\newline

Currently, Control Barrier Functions (CBF) are a popular technique to guarantee safety through the definition of forward invariant safe sets. Once a system is inside the safe set, future system states remain inside the safe set if the control inputs adhere to the CBF. Several methods have formulated a CBF for guaranteed collision avoidance in Model Predictive Control (MPC), using the Minkowski difference~\cite{chen2025} or a convex optimisation problem~\cite{dcbf} for polytopes, or via polynomial approximations of polygonal shapes~\cite{peng2023}.



\subsection{Spline Parameterisations}
A subset of direct collocation methods parametrise the states and/or controls as splines over a single interval or over the full horizon. The distinct advantage of this approach lies in the convex hull property of splines. This property ensures that if the coefficients of a spline satisfy a set of bounds, the entire continuous trajectory satisfies those bounds. It effectively eliminates the risk of constraint violation between time steps, while simultaneously allowing for coarse discretisation. Mercy et al.~\cite{Mercy2017} employ a full-horizon B-spline parameterisation for differentially flat systems and incorporate obstacle avoidance using the concept of separating hyperplanes. Vandewal et al.~\cite{Vandewal2020} extend~\cite{Mercy2017} to non-differentially flat systems with Bernstein basis functions and link the Bernstein coefficients to the Runge-Kutta integration scheme. Their method allows polynomial parametrisation over a single control interval instead of over the full horizon as~\cite{Mercy2017}. Both methods consider the hyperplane parameters as decision variables within the OCP. While this provides more numerical freedom to the optimisation solver, it introduces extra optimisation variables and non-convex, bi-linear constraints, rendering the OCP harder and more costly to solve.

\subsection{Contributions}
This paper presents a novel approach that aims to alleviate this computational bottleneck, applied to the domain of time-optimal motion planning. The determination of the separating hyperplanes is decoupled from the OCP, and instead executed using Least-Squares Support Vector Machine classification problems. It reduces the complexity of the optimisation problem while maintaining the rigorous safety guarantees provided by spline parameterisation. This work extends the results in~\cite{Dirckx2025} to the domain of spline-based motion planning and further improving its computational complexity by adopting a smooth minimum approximation. Section~\ref{sec:splines} firstly covers the Bernstein collocation method to create piecewise polynomial state trajectories and guarantee continuous collision avoidance. Section~\ref{sec:bilevel} introduces the transformation of the decoupled method to spline-based motion planning and lastly, Section~\ref{sec:experiments} details the experimental results and benchmark.

%% file: Sections/splines.tex
\label{sec:splines}
Consider a system with states $\vect{x} \in \mathcal{X} \subseteq \mathbb{R}^{l}$ and $\vect{u} \in \mathcal{U} \subseteq \mathbb{R}^{q}$, described by a continuous-time dynamical model: 
\begin{equation}
    \dot{\vect{x}} = \vect{f}(\vect{x}, \vect{u}, t),
\end{equation}
where $\vect{f}$ is a continuous function. In time-optimal motion planning, the goal is to reach a specific target state $\vect{x}_T$ from an initial state $\vect{x}_0$ in the shortest time $T$ possible. Additionally, the system often has to adhere to bounds on its states and controls such as velocity or acceleration limits, detailed as:
\vspace{-1ex}
\begin{align}
    &\vect{\underline{x}} \leq \vect{x} \leq \vect{\bar{x}},\\
    &\vect{\underline{u}} \leq \vect{u} \leq \vect{\bar{u}}.
\end{align}

\subsection{Collision Avoidance between Polytopes}
In this work, the robot's geometry is a circle and each of the $N_o$ obstacles is a polytope. Suppose each obstacle is described by a set $\mathcal{O} = \Set{\vect{v}_{j}}{\vect{v}_{j} \in V_{o}}$ where $V_{o}$ contains all the vertices of that obstacle's polytope. The separating hyperplane theorem expresses that the robot and a single obstacle do not overlap if there exists a hyperplane such that:
\begin{align}
    \boldsymbol{w}_l^{\top}\boldsymbol{v}_r(\vect{x}) +  b_l &\geq \epsilon + r_{v}, \label{eq:hyp_rob}\\
    \boldsymbol{w}_l^{\top}\boldsymbol{v}_{j} +  b_l &< 0,
\end{align}
where $l \in [1, N_o]$, $\boldsymbol{w}_l \in \mathbb{R}^{n_y}$ is the normal vector of the hyperplane in $n_y$-dimensional space, $b_l \in \mathbb{R}$ is the hyperplane's offset, $\vect{v}_r$($\vect{x}$) the state-dependent robot centre, $r_{v}$ the radius of the sphere and $\epsilon$ is a safety margin. For each robot-obstacle pair, a set of parameters $\vect{w}_l$ and $b_l$ should exist to state that the robot is collision-free.

\subsection{Optimal Control for Time-Optimal Motion Planning}
All the aforementioned components together lead to the subsequent continuous-time OCP for time-optimal motion planning, with the hyperplane parameters for all robot-obstacle pairs considered as optimisation variables:

\begin{equation}
    \begin{aligned}{}
        \centering
        \underset{\substack{\vect{x}, \vect{u}, T, \\ 
        \vect{w}, \vect{b}}}{\textbf{minimise}} \quad & T + \alpha\int_{0}^{T}\ell(\vect{x}, \vect{u}, T)dt\\
        \textbf{subject to} \quad & \dot{\vect{x}} = \vect{f}(\vect{x}, \vect{u}, t), \\
        & \boldsymbol{w}_l^{\top}\boldsymbol{v}_r(\vect{x}) + b_l \geq \epsilon + r_{v}, \\
        & \boldsymbol{w}_l^{\top}\boldsymbol{v}_{j} + b_l \leq 0, \\
        & \vect{x}_{0}= \vect{x}_{\text{start}}, \; \vect{x}_{T} = \vect{x}_{\text{goal}}, \\
        & \vect{\underline{x}} \leq \vect{x} \leq \vect{\bar{x}},\\
        & \vect{\underline{u}} \leq \vect{u} \leq \vect{\bar{u}},
        \label{eq:ocp_motplan}
    \end{aligned}
\end{equation}
where $T$ is the total trajectory time, $\vect{w} = [\vect{w}_1, \ldots, \vect{w}_{N_o}]$ and $\vect{b} = [b_1, \ldots, b_{N_o}]$ contain the hyperplane parameters for each robot-obstacle pair. This problem is an infinite-dimensional OCP and needs to be transcribed to a finite-dimensional NLP and solved using numerical optimisation techniques. This work adopts the combination of multiple shooting and Bernstein polynomial parameterisation from Vandewal et al.~\cite{Vandewal2020} as it extends the applicability of spline-based motion planning, and thus the method in this paper, to non-differentially flat systems.

\subsection{Bernstein Parameterisation}
To discretise Eq.~\eqref{eq:ocp_motplan}, the time interval $[0, T]$ is divided into $N$ equally spaced intervals of duration $h$, assuming a constant control $\vect{u}_k$ over each interval. In multiple shooting, the state trajectory can be discretised over a single interval through explicit integration of the system dynamics $\dot{\vect{x}} = \vect{f}(\vect{x}, \vect{u}, t)$. The following Runge-Kutta (RK) integration scheme of order $s$ can be used to achieve this: 
\begin{align}
    &\vect{x}(t_{k+1}) = \vect{x}(t_{k}) + h\sum_{i=1}^{s}b_{i}k_{i} \quad \text{with} \\
    &k_{i} = \vect{f}({\vect{x}(t_{k}) + h\sum_{l=1}^{i - 1}e_{il}k_{l}}, \vect{u}(t_{k}), t_{k} + c_{i}h), 
\end{align}
where $b{i}, c_{i}, e_{il}$ are the coefficients of the specific RK integration scheme of order $s$. Similarly, the state trajectory $\vect{x}(t) \in \mathbb{R}^n$ can also be parameterised in each control interval $t \in [t_k, t_k + h]$ using a Bernstein polynomial basis up to order $s$:
\begin{equation}
    \vect{x}(t) = \sum_{i=0}^{s} \beta_i B_i^s(\tau), \quad \tau = \frac{t-t_k}{h},
    \label{eq:bernstein}
\end{equation}
where $B_i^s(\tau) = \binom{s}{i} \tau^i (1-\tau)^{s-i}$ are the Bernstein basis polynomials of degree $s$, $\beta_i$ are the Bernstein coefficients, and $\tau \in [0,1]$.  The monomial coefficients of the $s^\text{th}$-order RK-integrator are related to the Bernstein coefficients as:
\begin{equation}
    \beta_i = \sum_{j=0}^i \frac{\binom{i}{j}}{\binom{s}{j}} b_j.
\end{equation}
Expressing the state trajectory in the Bernstein basis allows us to exploit the convex hull property over a single control interval. This property states that the spline is always fully contained in the control polygon formed by the spline coefficients~\cite{deboor1978}. By imposing constraints on these coefficients, the control polygon and thus the full spline satisfies these constraints, effectively guaranteeing continuous constraint satisfaction. For this property to hold, the constraints must be polynomial in state $x$ such that the overall constraint can again be expressed as a Bernstein basis and the convex hull property can be applied. \\

To promote more freedom for the robot to avoid an obstacle~\cite{Mercy2017}, the hyperplane $\vect{w}$ and $b$ are also parameterised over each interval $[t_{k}, t_{k+1}]$ using the Bernstein basis as
\begin{align}
    \boldsymbol{w}(t) &= \sum_{i=0}^{m} \beta_{w, i} B_i^m(\tau) \label{eq:w_spline}\\
    b(t) &= \sum_{i=0}^{m} \beta_{b, i} B_i^m(\tau), \label{eq:b_spline}
\end{align}
which allows the hyperplane to change over a single control interval. To limit the additional complexity in the OCP, only a linear change of the parameters as done in~\cite{Mercy2017} is allowed and thus $m=1$. Table~\ref{tab:lin_hyperplanes} discusses how the Bernstein coefficients of time interval $[t_{k}, t_{k+1}]$ are derived based on the hyperplane parameters at the start and end of that interval. These relations can be found by writing out Eq.~\eqref{eq:w_spline} and~\eqref{eq:b_spline} for $t_{k}$ and $t_{k+1}$ and solving for $\beta_0$ and $\beta_1$.

%% file: Sections/bilevel.tex
\label{sec:bilevel}
\subsection{Least-Squares Support Vector Machines}

\begin{table*}[t]
    \centering
    \tablefontsize
    \addtolength{\tabcolsep}{15.5pt}
    \begin{tabular}{ccc}
        \toprule[1.5pt]
        \textbf{SVM Vectors} & \textbf{Linear System} & \textbf{Quadratic Program} \\[0.2em]
        \midrule
        \makecell{
        $\vect{\Gamma} = [1, -\vect{1}_{N_o}]$ \\[0.5em]
         $\boldsymbol{Z} = [\boldsymbol{v}_{r}^{\top}\Gamma_1,\, \boldsymbol{v}_{1}^{\top}\Gamma_{2},\, \ldots,\, \boldsymbol{v}_{N_o}^{\top}\Gamma_{N_o + 1}]$} & $
            \begin{bmatrix}
                0 & -\boldsymbol{\Gamma}^{\top} \\[0.5em]
                \boldsymbol{\Gamma} & \boldsymbol{Z}\boldsymbol{Z}^{\top} + \frac{1}{\tau}\boldsymbol{I}\\
            \end{bmatrix}
            \begin{bmatrix}
                b \\
                \boldsymbol{\alpha}
            \end{bmatrix} = 
            \begin{bmatrix}
                0 \\
                \vect{1}
            \end{bmatrix}
        $
        &
        {$\!
            \begin{aligned}
                \underset{\boldsymbol{w}, b}{\textbf{min.}} \quad & \frac{1}{2} \boldsymbol{w}^{\top}\boldsymbol{w} \\
                \textbf{s.t.} \quad & \Gamma_{k}[\boldsymbol{w}^{\top}\boldsymbol{y}_{k}] + b \geq 1,\\
            \end{aligned}
        $} \\[1em]
        $\vect{\Gamma} \in \mathbb{R}^{N_d}$, $\boldsymbol{Z} \in \mathbb{R}^{N_d \times n_{y}}$ & $\boldsymbol{w} = \boldsymbol{Z}^{\top}\boldsymbol{\alpha}$ & 
        {$\!
            \vect{y}_k = \begin{cases}\vect{v}_r(\vect{x}), & \text{if $\Gamma_k = 1$}.\\
            \vect{v}_j, & \text{if $\Gamma_k = -1$}.
            \end{cases}
        $} \\
        \bottomrule[1.5pt]
    \end{tabular}
    \caption{Overview of the linear system and quadratic program used to compute a hyperplane between robot $\mathcal{R}$ and obstacle $\mathcal{O}$, with $N_{d} = N_{o} + 1$. $\vect{\alpha}$ is linearly related to the classification error of the hyperplane classifier, $\vect{\Gamma}$ represents the labels of the robot and obstacle class.}
    \label{tab:ls_qp_svm}
\end{table*}

Previous works have considered the normal vectors $\vect{w}$ and offsets $\vect{b}$ as additional optimisation variables inside an OCP~\cite{Vandewal2020,Mercy2017}. However, this introduces non-convex, bi-linear constraints and extra variables, making the OCP harder and more costly to solve. Conversely, the hyperplanes can be computed in a separate step between the iterations of the optimisation solver by solving a linear system (LS) or an inequality-constrained quadratic program (QP)~\cite{Dirckx2025}. The specific formulations of the LS and QP are detailed in Table~\ref{tab:ls_qp_svm}. \\

After solving either the LS or the QP, both $\boldsymbol{w}$ and $b$ are embedded as parameters into the OCP. The LS can be efficiently solved through the Cholesky decomposition of its Schur complement. Through decoupling the hyperplanes from the OCP in an environment with $M$ obstacles, the amount of variables and constraints are reduced by respectively $M(n_y + 1)$ and $MN_o$. This effect is detailed below where the optimisation variables are indicated in green. Since the second constraint on the left, concerning the obstacle and the hyperplane, does not contain any optimisation variables after decoupling, it must be removed from the OCP.
\begin{equation*}
\left.
\begin{aligned}
            \green{\boldsymbol{w}_l}^{\top}\boldsymbol{v}_r(\green{\vect{x}}) +  \green{b_l} &\geq \epsilon + r_{v}\\
            \green{\boldsymbol{w}_l}^{\top}\boldsymbol{v}_{j} +  \green{b_l} &< 0\\
\end{aligned}
\quad
\right\}
\longrightarrow
\quad
\boldsymbol{w}_l^{\top}\boldsymbol{v}_r(\green{\vect{x}}) +  b_l \geq \epsilon + r_{v}.
\end{equation*}
Since the computational cost to solve the NLP associated with Eq.~\eqref{eq:ocp_motplan} generally scales cubicly with the amount of constraints and variables~\cite{Waechter2005}, this leads to a significant cost reduction in obstacle-rich environments. Furthermore, the separating hyperplane constraints in Eq.~\eqref{eq:ocp_motplan} are non-convex if the hyperplane parameters are considered as optimisation variables. By decoupling the hyperplane computation from the OCP, the variables are thus eliminated and the corresponding constraints are linearised, which renders the OCP easier to solve. 




\subsection{Two-step Filter Procedure}

\begin{table}[t]
    \centering
    \tablefontsize
    \addtolength{\tabcolsep}{13.5pt}
    \begin{tabular}{ccc}
        \toprule[1.5pt]
         & $\vect{w}$ & $\vect{b}$ \\
        \midrule
        $\beta_0$ & $\vect{w}(t_{k})$ & $\vect{b}(t_{k})$ \\
        $\beta_1$ & $\vect{w}(t_{k+1}) - \vect{w}(t_k)$ & $\vect{b}(t_{k+1}) - \vect{b}(t_k)$ \\
        \bottomrule[1.5pt]
    \end{tabular}
    \caption{The relation between coefficients $\beta_0$ and $\beta_1$ of $\vect{w}$ and $b$ in terms of the time instances at which the hyperplanes are computed in the decoupled approach.}
    \label{tab:lin_hyperplanes}
\end{table}
If the hyperplane parameters are optimisation variables in Eq.~\eqref{eq:ocp_motplan}, they can change during iterations of the solver and result in a faster trajectory. To effectively also allow hyperplanes to change in the decoupled approach, the two-set filter procedure as in~\cite{Dirckx2025} is transformed to the case of continuous collision avoidance. The specific details related to this new procedure are detailed in Algorithm~\ref{alg:update_svm} and~\ref{alg:compute_svm}.

Firstly, Alg.~\ref{alg:update_svm} performs a check on whether or not the robot is in collision on all interval boundaries. Where the robot is in collision, all hyperplanes are updated. Secondly, if the robot is out of collision on all interval boundaries, the intervals where the robot is still in collision within the interval are identified. For all intervals where the robot is out of collision, the QP is solved to exploit the sparsity of its solution in terms of the hyperplane parameters.


\RestyleAlgo{ruled}
\SetKwComment{Comment}{/* }{ */}
\begin{algorithm}[h]
    \caption{Inter-iteration hyperplane computation - $\texttt{update}$}\label{alg:update_svm}
    \KwData{Current solution $\boldsymbol{x}_{v}$, set $R$ and $O$, $\boldsymbol{w}_{v, *}, \, w_{b, v, *}$}
    $\texttt{bounds\_coll} \gets \texttt{check\_bounds}(\boldsymbol{x}_{v}, R, O)$\\
    $\texttt{segs\_coll} \gets \texttt{check\_segments}(\vect{\beta}_{x, v}, R, O)$\\
    \eIf{$\texttt{bounds\_coll}$}
    {
    \For{$i, k \in bounds\_coll$}{
    
        $\boldsymbol{w}_{v + 1, i, k} \gets \texttt{solve}(\vect{v}_{r, i} \, ; \, V_{o, k} \, ; \, \boldsymbol{w}_{v, *} \, ; \texttt{True})$ \\
        $b_{v + 1, *} \gets \texttt{vertex}(\boldsymbol{w}_{v + 1, *} \, ; \, \vect{v}_{r, i} \, ; \, V_{o, k})$
    }
    }{
        \For{$i, k \in segs\_coll$}{
            $\boldsymbol{w}_{v + 1, i, k} \gets \texttt{solve}(\vect{v}_{r, i} \, ; \, V_{o, k} \, ; \, \boldsymbol{w}_{v, *} \, ; \texttt{True})$ \\
            $b_{v + 1, *} \gets \texttt{vertex}(\boldsymbol{w}_{v + 1, *} \, ; \, \vect{v}_{r, i} \, ; \, V_{o, k})$
        } 
    }
\end{algorithm}
\vspace{-5ex}
\begin{algorithm}[h]
\caption{Computing the hyperplane normal $\boldsymbol{w}$ - $\texttt{solve}$}\label{alg:compute_svm}
\KwData{Robot center $\vect{v}_{r, i}$, obstacle vertices $V_{l}$, normal $\boldsymbol{w}_{v, i, k}$, $\texttt{LS}$}
\tcc{Table~\ref{tab:ls_qp_svm}}
$\boldsymbol{w}_{new} \gets \texttt{solve\_svm}(\vect{v}_{r, i} \, ; \, V_{o, k}  \, ; \, \texttt{LS})$ \\
$\theta_{tr} = \texttt{trust\_region}(\boldsymbol{w}_{v, i, k} \, ; \, \boldsymbol{w}_{new})$\;
\If{$\theta_{tr} \leq \bar{\theta}$}{
    $\boldsymbol{w}_{v + 1, i, j, l} \gets \boldsymbol{w}_{new}$
}
\end{algorithm}




%% file: Sections/experiments.tex
\begin{figure}
    \centering
    \includegraphics[width=0.65\columnwidth]{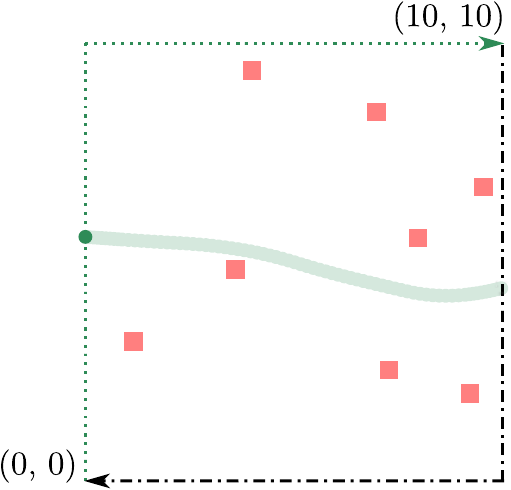}
    \caption{Illustration of a validation environment with ten obstacles, a collision-free trajectory through it and the variation in start and end position to generate the full experimental data set.}
    \label{fig:env_illust}
\end{figure}

\label{sec:experiments}
The proposed contributions are validated on a two-dimensional motion planning case, as illustrated in Figure~\ref{fig:env_illust}. A robot with circular footprint has to avoid an increasing amount (up to 10) of rectangular obstacles. For each environment, as visualised in Figure~\ref{fig:env_illust}, the start position varies clockwise from $[0, 0]$ to $[10, 10]$ while the end position varies counter-clockwise. This allows us to cover many different trajectories for a single environment. Each OCP is initialised with a straight line between start and end position, leading to some infeasible initialisations where the robot is in collision with an obstacle somewhere along the straight line. The robot is modelled according to the unicycle model $\dot{\boldsymbol{x}} = \vect{f}(\boldsymbol{x}, [v, \omega]) = [v\cos(\theta), v\sin(\theta), \omega]^{\top}$. The state vector $\boldsymbol{x} = [x, y, \theta]^{\top}$ contains the position $\boldsymbol{p}_r = [x, y]^{\top}$ of the AMR and orientation $\theta$. Forward velocity $v$ and steering angle $\delta$ are the robot's control inputs. The initial guess is constructed using a straight line in Cartesian space between start and end point, and by solving the LS and the QP on the infeasible and feasible points along that initial line. \\

The performance of a decoupled spline-based motion planner is studied on several metrics: computational efficiency (in total and per iteration), success rates in a cluttered environment and time-optimality of the final solution. The proposed method is benchmarked against the method in~\cite{Vandewal2020}, termed LSR. The OCP is formulated in CasADi~\cite{Andersson2018} and solved with IPOPT, using the linear solver \texttt{ma27}~\cite{hsl}. The linear solver `lapacklu' is used to solve the LS while ProxQP is used to solve the QP. All experiments are conducted on a portable computer on an Intel® Core™ i7-10810U processor with twelve cores at 1.10GHz and 31.1 GB RAM. Ten and twenty degrees are taken as trust-region boundary $\bar{\theta}$ in the first and second use of the \texttt{solve} function in Alg.~\ref{alg:compute_svm}, respectively.\\

\begin{table*}[h]
    \centering
    \addtolength{\tabcolsep}{10.25pt}
    \begin{tabular}{crcccccccc}
        \toprule[1.25pt]
        \textbf{Metrics} & \textbf{Method} & \multicolumn{8}{c}{$N_o$}\\
        \midrule[0.25pt]
        \addlinespace[1ex]
         & & 1 & 2 & 3 & 4 & 5 & 6 & 7 & 8 \\ 
        \midrule[0.25pt]
        \addlinespace[1ex]
        $t_{\text{LS}}$ (ms) & SVM & 0.42 & 0.64 & 0.45 & 0.79 & 0.86 & 0.74 & 0.61 & 0.77 \\ 
        \midrule[0.25pt]
        \addlinespace[1ex]
        $t_{\text{QP}}$ (ms) & SVM & 1.31 & 1.01 & 1.77 & 1.90 & 2.36 & 1.86 & 2.29 & 2.75 \\ 
        \midrule[0.25pt]
        \addlinespace[1ex]
        \multirow{2}{*}{$t_{\text{wall}}$ (ms)} & SVM & \textbf{25.12} & \textbf{31.55} & \textbf{40.12} & \textbf{47.42} & \textbf{58.02} & \textbf{65.96} & \textbf{70.77} & \textbf{83.70} \\ 
         & LSR & 35.82 & 50.28 & 77.10 & 107.12 & 125.66 & 144.58 & 168.61 & 204.46 \\ 
        \midrule[0.25pt]
        \addlinespace[1ex]
        \multirow{2}{*}{\# I} & SVM & \textbf{12} & \textbf{12} & \textbf{13} & \textbf{15} & \textbf{16} & \textbf{18} & \textbf{18} & \textbf{20} \\ 
         & LSR & 15 & 15 & 18 & 21 & 23 & 23 & 24 & 26 \\ 
        \midrule[0.25pt]
        \addlinespace[1ex]
        $\epsilon_{\text{move}}$ (\%) \makecell{med. \\ std.} & SVM & \makecell{0.0 \\  1.56} & \makecell{0.0 \\  5.81} & \makecell{0.0 \\  4.11} & \makecell{0.05 \\  13.13} & \makecell{0.13 \\  9.91} & \makecell{0.18 \\  12.00} & \makecell{0.32 \\  20.66} & \makecell{0.44 \\  24.32} \\ 
        \midrule[0.25pt]
        \addlinespace[1ex]
        \multirow{2}{*}{Success (\%)} & SVM & \textbf{100} & 97.5 & 97.5 & 97.0 & 93.0 & 92.0 & 87.5 & 87.0 \\ 
         & LSR & 93.5 & \textbf{100} & \textbf{100} & \textbf{100} & \textbf{100} & \textbf{100} & \textbf{100} & \textbf{99.0} \\ 
        \midrule[1.25pt]
        \addlinespace[1ex]
        Reduction (\%) & & 29.87 & 37.24 & 47.96 & 55.74 & 53.83 & 54.38 & 58.03 & 59.06 \\ 
        \bottomrule[1.25pt]
    \end{tabular}
    \caption{Summary of the results up to eight obstacles: (1, 2) the maximum computation time added by solving the LS or QP, (3) the median computation time to solve Eq.~\eqref{eq:ocp_motplan}, (4) the median number of iterations to solve Eq.~\eqref{eq:ocp_motplan}, (5) the median and standard deviation of the error on trajectory time, (6) the success rate and (7) the reduction in computation time from LSR to the decoupled method. All reported maximum and median numbers are computed over all successful trajectories per $N_\text{o}$.}
    \label{tab:splines_metrics}
\end{table*}


\begin{figure*}
    \centering
    \includegraphics[width=0.95\textwidth]{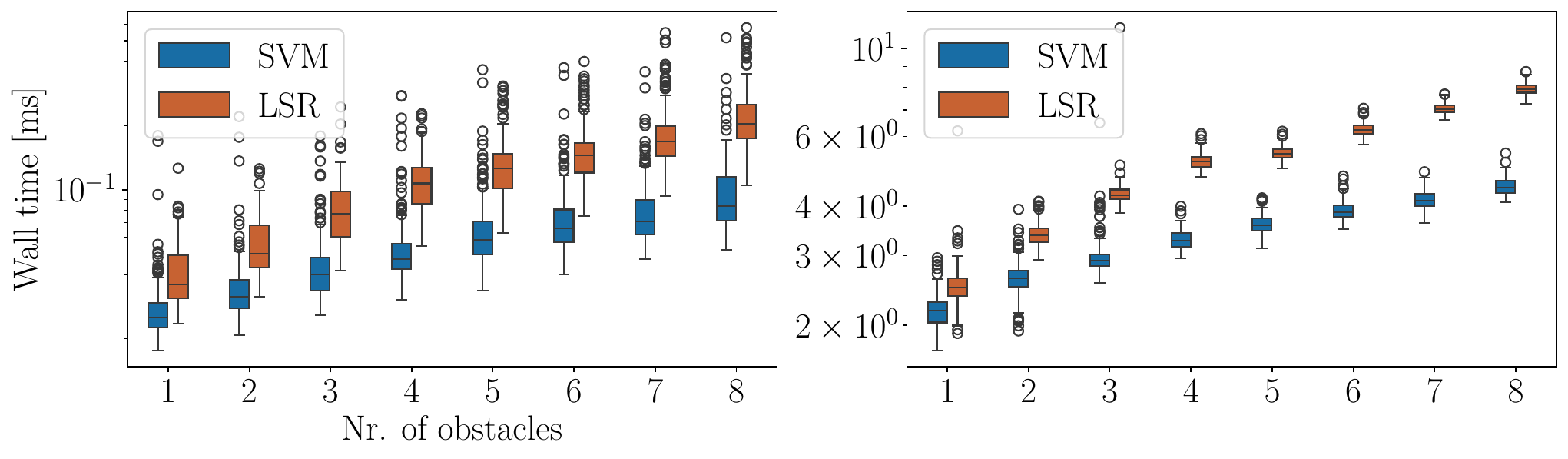}
    \caption{Results of the comparison between the coupled and decoupled approach in terms of (top) computation time per iteration and (bottom) total computation time to solve Eq.~\eqref{eq:ocp_motplan} per $N_{\text{o}}$.}
    \label{fig:splines_wall_time}
\end{figure*}

Firstly, Figure~\ref{fig:splines_wall_time} reports the total wall time and wall time per iteration necessary to solve Eq.~\eqref{eq:ocp_motplan} up to eight obstacles. Secondly, Table~\ref{fig:splines_wall_time} reports the maximum wall time added to solve the LS or QP while solving Eq.~\eqref{eq:ocp_motplan}, the median total wall time $t_{\text{wall}}$ and number of iterations required to solve the OCP, the median and standard deviation of the error $\epsilon_{\text{move}}$ in motion time $T$ where LSR is taken as baseline, the success rate over the full set of 200 trajectories and the reduction in $t_{\text{wall}}$ achieved by adopting the decoupled approach compared to LSR. They show the improvement in computational time by adopting the decoupled approach, certainly for more than one obstacle. A reduction up to a factor 2.44 (59.06\%) can be achieved by eliminating the hyperplanes as optimisation variables for up to eight obstacles. As seen in Table~\ref{tab:splines_metrics}, the median optimality in terms of motion time increases slightly but this increase is negligible relative to the gain in computation time. Most importantly, this reduction is achieved while maintaining the continuous constraint satisfaction guarantees from LSR. \\

However, the decoupled approach introduces a non-trivial trade-off between computational efficiency and robustness. As the obstacle density increases, the method exhibits a noticeable degradation in success rates and a widening optimality gap regarding trajectory time. This limitation stems from two factors: the conservative free-space approximation yielded by the LS formulation and, more critically, the use of a zero-order approximation for the hyperplane constraints. Because the numerical solver lacks first-order sensitivity information regarding how the hyperplanes evolve relative to the trajectory, it is prone to getting stuck in sub-optimal local minima.



%% file: Sections/conclusion.tex
\label{sec:conclusion}

This work presented a pragmatic advancement in the domain of autonomous spline-based motion planning for non-differentially flat systems. By decoupling the determination of separating hyperplanes from the primary OCP and reformulating them as classification problems (either a LS or QP), this work effectively transform non-convex, bi-linear constraints into linear ones. Furthermore, by eliminating hyperplane parameters as decision variables, the OCP's dimensions are reduced, making it significantly computationally cheaper to solve. Experimental validation demonstrates a compelling reduction in wall time compared to a state-of-the-art method, called Local Spline Relaxation (LSR), up to nearly 60\% in environments with eight obstacles. Crucially, this efficiency is achieved without sacrificing the continuous safety guarantees provided by the Bernstein basis parameterisation. \\

Future extensions to this work include adopting structure-exploiting algorithms and state-of-the-art solvers such as BLASFEO~\cite{Frison2018} (LS) and HPIPM~\cite{hpipm} (QP). This would further decrease the computation times in Table~\ref{tab:splines_metrics} and increase scalability of this approach to obstacle geometries with more vertices. Moreover, introducing first-order derivative information about the solution of the classification problem to the trajectory is a necessary step to reduce the optimality gap and increase the success rate.